\documentclass[a4paper, 10pt, conference]{ieeeconf}      
\usepackage{FG2023}

\FGfinalcopy 

\usepackage{graphicx}
\usepackage{hyperref}
\usepackage{makecell}
\usepackage{multirow}
\usepackage{physics}

\IEEEoverridecommandlockouts                              
\def\FGPaperID{****} 

\title{\LARGE \bf
Mobile Keystroke Biometrics Using Transformers}

\author{\parbox{16cm}{\centering
    {\large Giuseppe Stragapede$^1$, Paula Delgado-Santos$^{2, 1}$}, Ruben Tolosana$^1$, Ruben Vera-Rodriguez$^1$, Richard Guest$^2$, Aythami Morales$^1$\\
    {\normalsize
    $^1$ Biometrics and Data Pattern Analytics Lab, Universidad Autonoma de Madrid, Spain \\
    $^2$ School of Engineering, University of Kent, UK}}
}

\begin{document}

\ifFGfinal
\thispagestyle{empty}
\pagestyle{empty}
\else
\author{Anonymous FG2023 submission\\ Paper ID \FGPaperID \\}
\pagestyle{plain}
\fi
\maketitle

\begin{abstract}
Among user authentication methods, behavioural biometrics has proven to be effective against identity theft as well as user-friendly and unobtrusive. One of the most popular traits in the literature is keystroke dynamics due to the large deployment of computers and mobile devices in our society. This paper focuses on improving keystroke biometric systems on the free-text scenario. This scenario is characterised as very challenging due to the uncontrolled text conditions, the influence of the user's emotional and physical state, and the in-use application. To overcome these drawbacks, methods based on deep learning such as Convolutional Neural Networks (CNNs) and Recurrent Neural Networks (RNNs) have been proposed in the literature, outperforming traditional machine learning methods. However, these architectures still have aspects that need to be reviewed and improved. To the best of our knowledge, this is the first study that proposes keystroke biometric systems based on Transformers. The proposed Transformer architecture has achieved Equal Error Rate (EER) values of 3.84\% in the popular Aalto mobile keystroke database using only 5 enrolment sessions, outperforming by a large margin other state-of-the-art approaches in the literature.

\end{abstract}


\section{Introduction}
\label{sec:Introduction}
Due to the increasing number of online transactions and fraudulent activities in sectors such as Banking, Financial Services and Insurance (BFSI), healthcare, e-commerce, and government, among many others, the demand and the investments for more secure and reliable digital authentication methods are rising \cite{bbreport}. Such trend is particularly relevant with regard to mobile devices, given their popularity. 
\par Recent authentication methods propose to increase the security through an additional transparent layer based on the user's behavioural biometric information\footnote{In contrast to \textit{physiological} biometrics, which pertains to the biological characteristics of an individual, such as face or fingerprint, all means that enable or contribute to differentiating between individuals throughout the way they perform activities are labelled as \textit{behavioural}, i.e., gait, keystroke dynamics, handwritten signature, etc.}, overcoming potential identity theft in a user-friendly and continuous way \cite{ISO2018, 7503170}.
Among the different behavioural biometric traits, keystroke dynamics is one of the most popular authentication methods in the literature \cite{acien2021typenet, Maiorana2021}. The information considered is the timestamp of the actions of pressing and releasing a key, together with the information of the key typed.


\par Keystroke biometric systems are typically divided into two groups \cite{mondal2017study}: \textit{fixed-text}, where the keystroke sequence typed by the user is prefixed, such as a username or password, and \textit{free-text}, where the keystroke sequence is arbitrary, such as writing an email. In the latter, typing errors are common, and the keystroke sequences between the enrolment and test samples are different, contrary to fixed-test scenarios. Consequently, the performance achieved with free-text keystroke systems are traditionally lower than their fixed-text counterparts due to the higher intra-subject variability and complexity of the task.


\par In addition, focusing on keystroke biometrics on mobile scenarios, many challenges must be considered to develop robust authentication systems. In this particular scenario, keystroke is typically acquired under uncontrolled circumstances, which can be affected by the user's activity, body position, emotional state, and the acquisition device \cite{Maiorana2021, TEH2016210}. The performance might also be affected if the same subject is able to speak different languages \cite{Abuhamad2021}.

\par In the present work, we explore and propose a Transformer architecture to overcome the challenges commented before and improve the authentication performance of free-text keystroke biometric systems on mobile scenarios. Originally proposed in \cite{vaswani2017attention}, Transformers are defined by an encoder-decoder architecture. They have quickly gained the attention of the scientific community given their ability to model a number of different processes, in fields such as computer vision, machine translation, reinforcement learning, time-series analysis for classification and prediction, etc. \cite{tay2020efficient}. These new architectures have shown several advantages compared with Convolutional Neural Network (CNNs) and Recurrent Neural Networks (RNNs): \textit{(i)} they process all sequences in parallel being feed-forward models, \textit{(ii)} they operate over long-distance sequences applying self-attention mechanism, \textit{(iii)} they undergo a more efficient training allowing the process of all samples in one batch, and \textit{(iv)} they attend to all of the previous sequence at the same time without the need to summarise the seen information \cite{vaswani2017attention}.
\par 

\par In summary, the main contributions of this work can be listed as follows:

\begin{figure*}[t]
    \centering
     \includegraphics[trim={0cm 0cm 0 0cm}, width=0.9\linewidth]{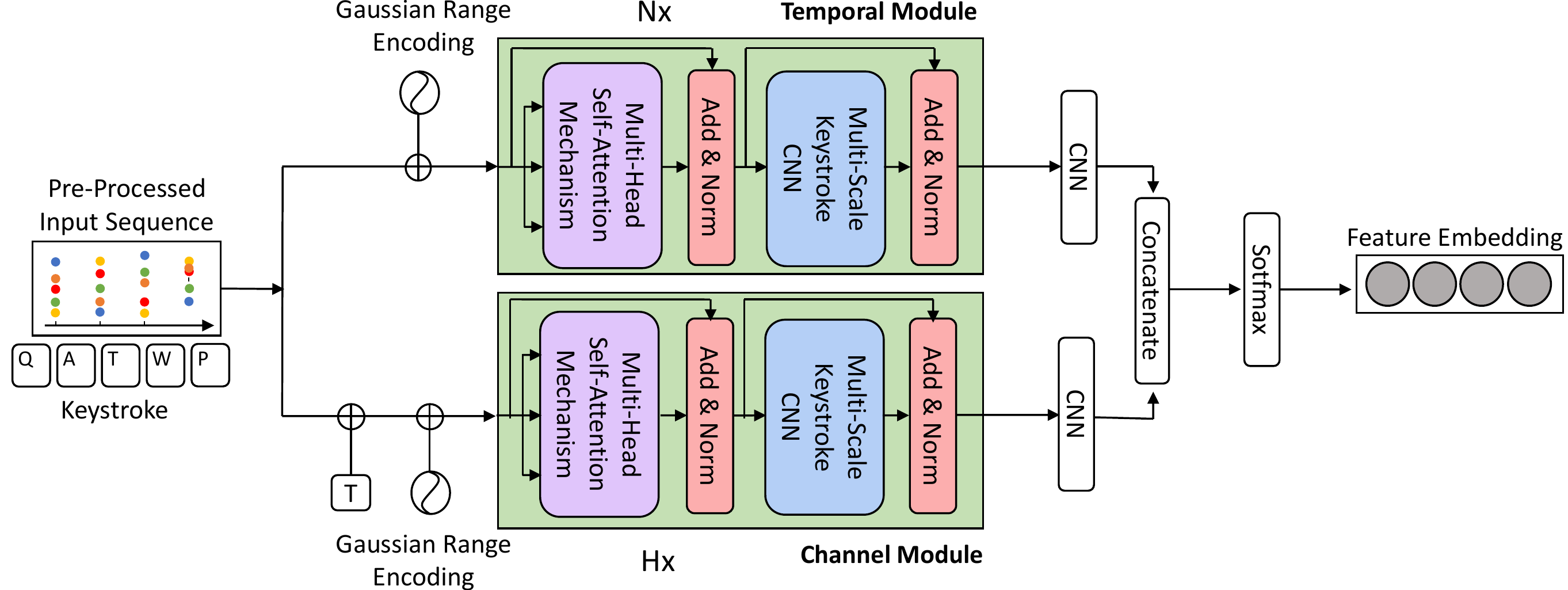}
    \caption{Graphical representation of the proposed Transformer architecture. T: Transposition; N, H: Number of layers of each of the modules.}
    \label{fig:TransformerArch}
\end{figure*}

\begin{itemize}
    \item Novel keystroke verification system for the challenging free-text mobile scenario based on Transformers. Fig. \ref{fig:TransformerArch} provides a graphical representation of the proposed Transformer architecture. To the best of our knowledge, this is the first study that explores Transformers for keystroke biometrics.
    \item Comparison of the proposed Transformer with previous approaches in the literature using the popular and public Aalto mobile keystroke database \cite{palin2019people}. The proposed approach achieves Equal Error Rate (EER) values of 3.84\% using only 5 enrolment sessions, outperforming by a large margin previous state-of-the-art approaches.
    \item We make our proposed approach and experimental framework available to the research community in order to advance the state of the art of keystroke biometrics in free-text mobile scenarios\footnote{\texttt{\url{https://github.com/BiDAlab/TypeFormer}}}.
\end{itemize}

\par The remainder of the paper is organised as follows: Sec. \ref{sec:TheConsideredDatabase} summarises the Aalto mobile keystroke database. Sec. \ref{sec:ProposedSystem} describes the feature extraction process and the proposed Transformer architecture. Then, in Sec. \ref{sec:experimentalprotocol}, we present a detailed description of the experimental setup. Sec. \ref{sec:experimentalresults} contains the experimental results of the proposed approach, and the comparison with the state of the art. Finally, Sec. \ref{sec:conclusions} draws some conclusions and future research lines.


\section{The Aalto Mobile Keystroke Database}
\label{sec:TheConsideredDatabase}
The Aalto mobile keystroke database comprises free-text keystroke dynamics data from around 260,000 subjects \cite{palin2019people}. A mobile web application was implemented for the data acquisition, in a totally unsupervised way. The subjects were asked to read English sentences, and to type them as rapidly and accurately as possible in their own smartphones. The provided sentences were randomly withdrawn from a set of 1,525 sentences obtained from the Enron mobile mail \cite{10.1145/2037373.2037418} and the Gigaword Newswire corpora \cite{gigaword}. The requisite for each of the sentences was containing at least 3 words and at most 70 characters. Around 68\% of the participating subjects were English native speakers. The raw data recorded consists in the acquisition of key press and key release events from the browser, with the resolution of 1ms. In the present work, we select all subjects (62,454) that completed at least 15 acquisition sessions.

\section{Proposed System}
\label{sec:ProposedSystem}
\begin{figure}[tp]
    \centering
     \includegraphics[trim={0cm 0cm 0 0cm}, width=0.8\linewidth]{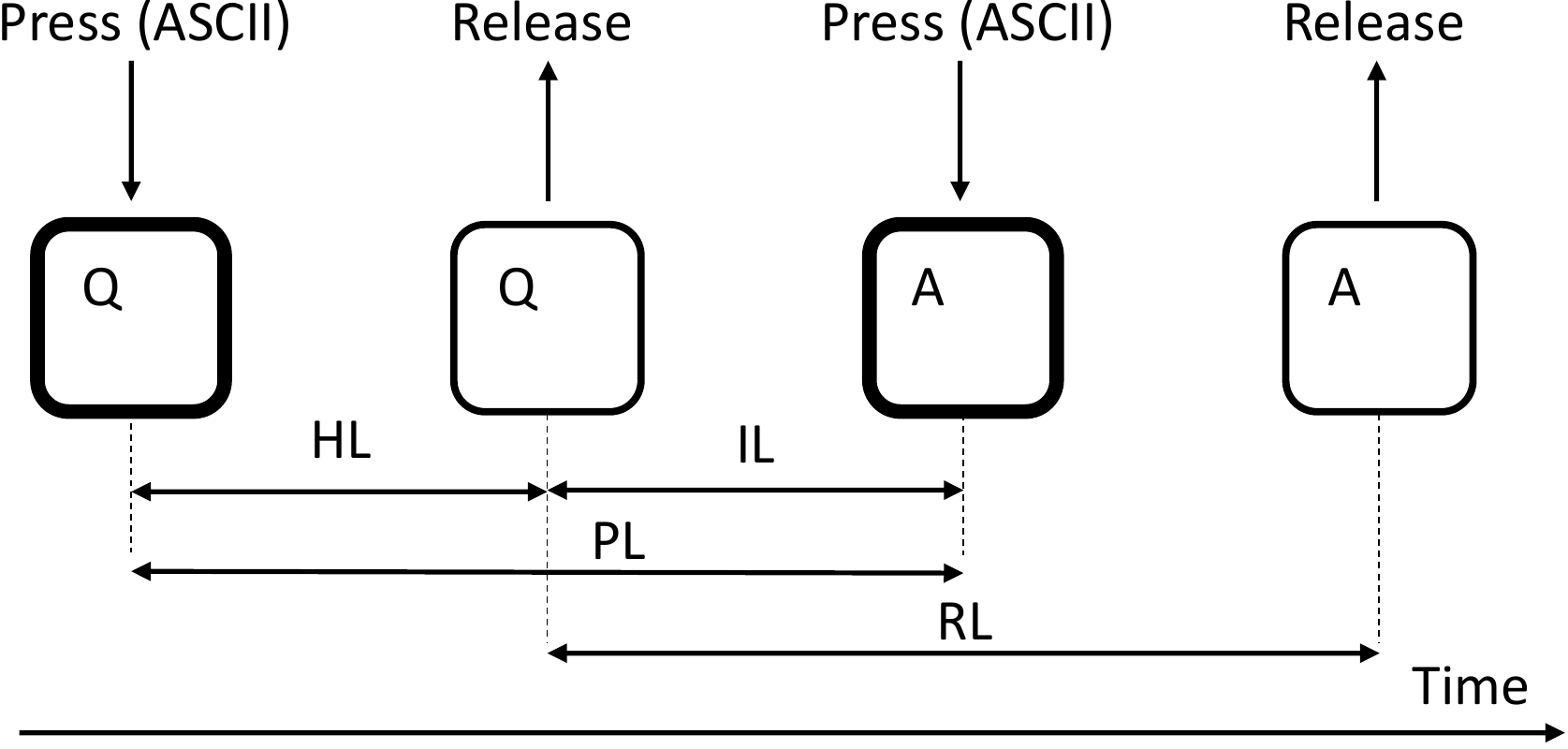}
    \caption{Example of the keystroke features extracted from the Aalto mobile keystroke database \cite{palin2019people}. HL: Hold Latency; IL: Inter-key Latency; PL: Press Latency; RL: Release Latency; ASCII: Key Pressed.}
    \label{fig:keystrokefeatures}
\end{figure}


This section provides the details of the proposed keystroke verification system for free-text mobile scenarios. 

\subsection{Feature Extraction}
\label{subsec:FeatureExtraction}
The raw data are pre-processed following the approach described in \cite{acien2021typenet}. Data consist in the timestamp of the action of pressing and releasing a key, together with the ASCII code typed. The set of 5 features reported below are extracted per each press-release action:
\begin{center}
[\textit{hold latency}, \textit{inter-key latency}, \textit{press latency}, \textit{release latency}, \textit{key pressed}]
\end{center}
The five considered features are illustrated in Fig. \ref{fig:keystrokefeatures}.
Since the length of the text considered in each of the acquisition sessions is not constant (free-text scenario), they are sliced or zero-padded to obtain a sequence of $L = 50$ samples, aiming to minimise the system input duration. The ASCII code (key pressed) is normalised in the $[0,1]$ interval. 


\subsection{Transformer Architecture}
\label{subsec:TransformerArchitecture}


\par Fig. \ref{fig:TransformerArch} provides a graphical representation of the proposed Transformer, based on an adaptation of the encoder part of the Vanilla Transformer \cite{vaswani2017attention}. The Vanilla Transformer was tested in several fields showing impressive results but needs some adaptations in order to be used in time sequences. Several researchers introduced new aspects such as reduced complexity, periodicity-based dependencies, or time-depending encoding \cite{tay2020efficient}. We describe next the key aspects of our proposed Transformer.

\begin{figure}[tp]
    \centering
     \includegraphics[trim={0cm 0cm 0 0cm}, width=0.7\linewidth]{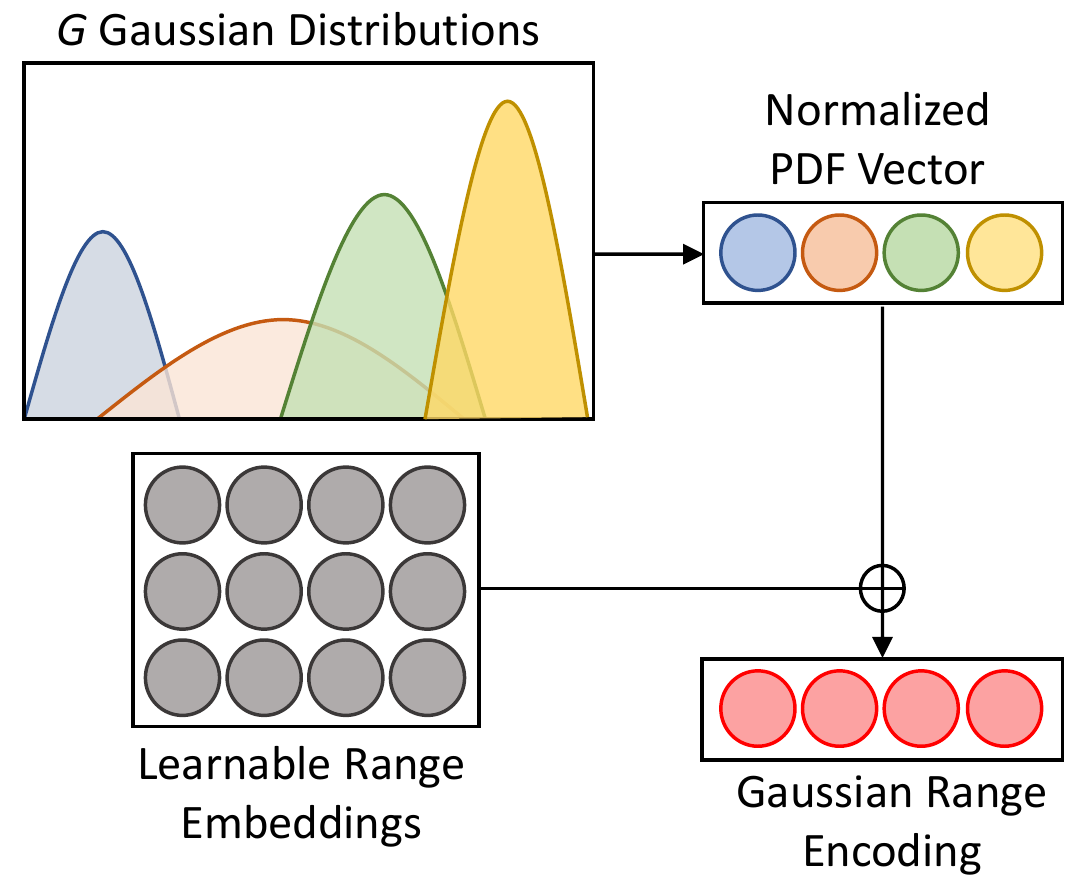}
    \caption{Graphical representation of the Gaussian range encoding. PDF: Probability Density Function.}
    \label{fig:GaussianRangeEncoding}
\end{figure}

The pre-processed input sequence $\vb{X}=(\vb{x}_0,\vb{x}_1,...,\vb{x}_l,...,\vb{x}_L)$ is introduced into the Transformer model. Adopting the idea presented in \cite{li2021two}, we have first changed the original positional encoding by a \textit{Gaussian range encoding}. Fig. \ref{fig:GaussianRangeEncoding} provides a graphical representation of the Gaussian range encoding. The pre-processed input sequence is modelled with $G$ Gaussian distributions where the Probability Density Function (PDF) vector is a L1-normalized vector of the Gaussian PDFs. Furthermore, more than one range can be used at the same time, obtaining a more complex context position of each sample compared with the original positional encoding. The global Gaussian range encoding is the pondered multiplication of the PDF vector in the different ranges. It is important to highlight that, contrary to \cite{li2021two}, we have considered the Gaussian range encoding in both branches of the Transformer (Temporal and Channel Modules).

After the Gaussian range encoding, the proposed Transformer changes the original layer considered in the Vanilla Transformer by the two different modules considered in \cite{li2021two}: \textit{(i)} Temporal Module, and \textit{(ii)} Channel Module. The Temporal Module extracts information from the original input sequence (temporal-over-channel features), while the Channel Module transposes the input sequence to extract channel-over-temporal features. The Temporal Module contains a stack of $N$ identical layers while the Channel Module contains a stack of $H$ identical layers. Each module comprises two sub-layers: \textit{(i)} a multi-head self-attention mechanism, and \textit{(ii)} a multi-scale keystroke CNN. Then, each sub-layer is followed by a residual connection and a layer normalisation (Add \& Norm in Fig. \ref{fig:TransformerArch}). We provide next the essential details of the \textit{multi-head self-attention mechanism} and the \textit{multi-scale keystroke CNN} sub-layers.

The \textit{multi-head self-attention mechanism} is responsible for linking each of the samples along the entire input sequence. The procedure extracts long-range dependencies without limiting the time window size. The output of the sub-layer is the weighted summation of the values $V$ in accordance with the dot-product of the queries $Q$ and the matching keys $K$ \cite{vaswani2017attention}. The output of the sub-layer is the concatenation of applying the attention mechanism to a $F$ independent heads.
The \textit{multi-scale keystroke CNN} comprises convolutional layers with ReLU activation and different kernel sizes. A batch normalisation and a dropout layer are introduced in between.

A convolutional block is placed after each module. The CNN features are then concatenated and introduced to a softmax layer. The feature embedding obtained is $\vb{P}=(\vb{p}_0,\vb{p}_1,...,\vb{p}_s,...,\vb{p}_S)$,  where $S$ is the number of output features. Finally, for the verification task considered in the present study, the feature embeddings of the enrolment and test samples are compared using the Euclidean distance. It is important to highlight that the architecture configuration and model hyperparameters of the proposed Transformer have been adapted to free-text keystroke verification systems on mobile devices. The specific details of the proposed Transformer are described in Sec. \ref{subsec:ModelHyperparameters}.



\section{Experimental Setup}
\label{sec:experimentalprotocol}
\begin{figure*}[tp]
    \centering
     \includegraphics[trim={0cm 0cm 0 0cm}, width=0.95\linewidth]{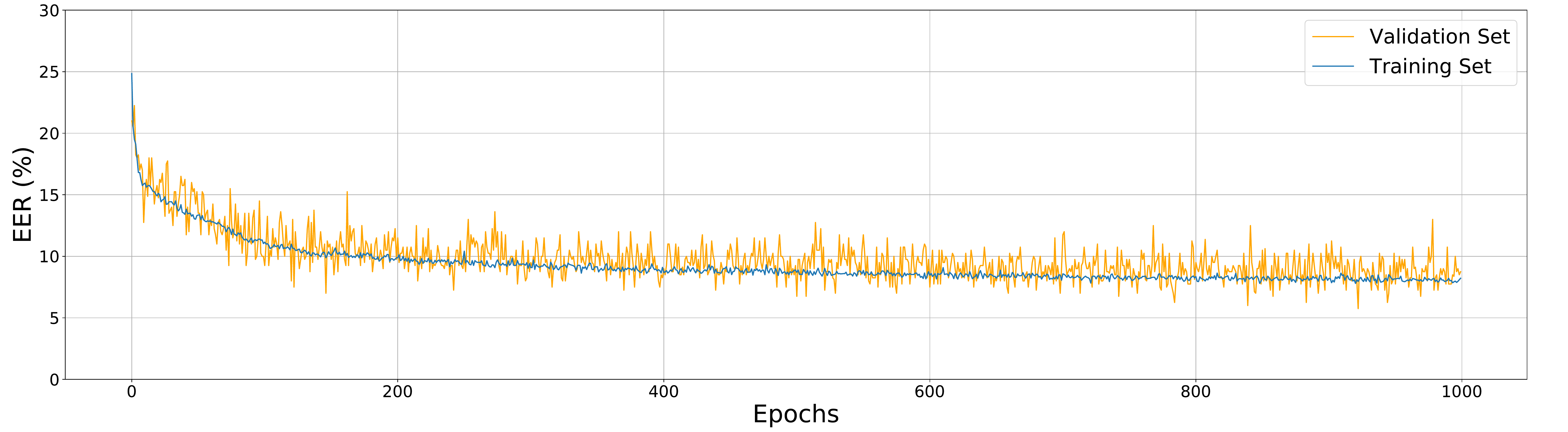}
    \caption{The EERs [\%] achieved on the training and validation sets at the end of each epoch of the training process are displayed above.}
    \label{fig:dev_eer}
\end{figure*}

\subsection{Transformer Hyperparameters}
\label{subsec:ModelHyperparameters}

This section describes the optimal hyperparameters of the proposed Transformer. These hyperparameters have been selected using the development experimental protocol described in the following section, Sec. \ref{subsec:ModelDevelopment}.

The Gaussian range encoding relies on $G = 20$ Gaussian distributions. The Temporal Module comprises $N = 10$ identical layers whereas the Channel Module comprises $H = 1$ layer. Regarding the multi-head self-attention sub-layer, $F = 10, 5$ heads are considered respectively for the Temporal and Channel Modules. In both modules, the multi-scale keystroke CNN comprises 3 convolutional layers with $L$ units each, ReLU activation functions, and kernel sizes 1, 3, and 5, respectively, followed by dropout layers with a rate of 0.1. Then, after the Temporal and Channel Modules, we consider 2 convolutional layers ($L$ units each, ReLU activation functions, and kernel sizes 128 and 32 respectively, followed by dropout layers with a rate of 0.5) with max-pooling, and a linear layer with softmax activation function. The size of the final feature embedding is $S = 64$.

\subsection{Model Development}
\label{subsec:ModelDevelopment}
We follow the public experimental protocol presented by Acien \textit{et al.} in \cite{acien2021typenet}, considering the same 30,000 subjects for training the models and 400 for validation. In total, 15 sessions per subject are considered. The proposed Transformer is implemented in \texttt{PyTorch}. Its training relies on a triplet loss function based on Euclidean distance with a margin $\alpha = 1.0$. Adam optimiser with default parameters and learning rate value of 0.001 is considered. We train the Transfomer for 1,000 epochs in total, considering 29 batches sized 1024 per epoch, i.e., 29,696 triplets. The selection of triplets takes place randomly with a uniform distribution. At the end of each epoch, the model is evaluated on the entire validation set, and when achieving a lower EER value, the corresponding model is saved. Fig. \ref{fig:dev_eer} provides a graphical representation of the training/validation results of the proposed Transformer along the number of epochs. In general, we can observe a smooth training curve in time.

\subsection{Model Evaluation}
\label{subsec:ModelEvaluation}
The best model selected in the development stage is finally evaluated using the public experimental protocol considered in \cite{acien2021typenet}. This evaluation consists of 1,000 unseen subjects, not considered in the development stage. In addition, from the total 15 sessions available per subject, we perform experiments using different configurations of enrolment sessions ($E = 1, 5, 10$), similar to \cite{acien2021typenet}, in order to assess the system adaptation to reduced availability of enrolment data. To obtain the genuine score distribution, for each user, we always consider the last 5 sessions for testing. Each of them is compared with the $E$ enrolment sessions obtaining $5 \times E$ scores, which are averaged over the $E$ enrolment sessions, leading to 5 final genuine scores per user. Regarding the impostor score distribution, we follow the same approach, but this time each user is compared with one test session of the remaining users, leading to 999 impostor scores per user.
Furthermore, it is important to highlight that from here two different approaches can be adopted to compute the EER: \textit{(i)} selecting an individual threshold for each of the enrolled subjects obtaining an EER per subject and then obtaining their average value as the final EER value; \textit{(ii)} selecting a unique threshold for all subjects. Their use depends on the particular application and scenario. The advantage of \textit{(i)} consists in a better performance in terms of EER, as a specific threshold of the system is adapted to each subject. This is the approach adopted in the comparison work \cite{acien2021typenet}. However, the drawback is related to the fact that a large impostor embedding distribution has to be compared with each of the individual subjects' feature embedding to tune each specific threshold. Such solution might be in conflict with the mobile environment resource constraints in terms of memory, or with privacy and security concerns. In fact, it has been shown that a significant amount of information can be obtained from data of mobile devices such as keystroke \cite{delgadosantos2021survey}. On the other hand, if the system is trained \textit{offline} on a large database, deploying the system with a fixed pre-determined unique threshold \textit{(ii)} is undoubtedly more convenient due to the described aspects. In the present work, both scenarios are considered in the experimental framework, naming them respectively ``Average" EER for case \textit{(i)}, and ``Global" EER for case \textit{(ii)}.

\section{Experimental Results}
\label{sec:experimentalresults}
\subsection{Comparison with the state of the art}

\begin{table}[t]
    \centering
    \caption{System performance comparison between the state-of-the-art TypeNet \cite{acien2021typenet} and the proposed Transformer.}
    \begin{tabular}{{|ccccc|}}
    \hline
    \multirow{2}{*}{\makecell{Number of \\ Enrolment \\ Sessions}} & 
    \multicolumn{2}{|c|}{\makecell{Average \\ EER (\%)}} & 
    \multicolumn{2}{c|}{\makecell{Global \\ EER (\%)}} \\
    \cline{2-5}
    & 
    \multicolumn{1}{|c|}{\makecell{\textbf{TypeNet} \\ \textbf{\cite{acien2021typenet}}}} &
    \multicolumn{1}{c|}{\makecell{\textbf{Proposed} \\ \textbf{Tranformer}}} &
    \multicolumn{1}{c|}{\makecell{\textbf{TypeNet} \\ \textbf{\cite{acien2021typenet}}}} &
    \multicolumn{1}{c|}{\makecell{\textbf{Proposed} \\ \textbf{Transformer}}}
    \\
    \hline
    \multicolumn{1}{|c}{1} &
    \multicolumn{1}{|c|}{12.60} & 
    \multicolumn{1}{c|}{\textbf{6.99}}  & 
    \multicolumn{1}{c|}{18.19} & 
    \multicolumn{1}{c|}{\textbf{10.68}} 
    \\ 
    \hline
    \multicolumn{1}{|c}{5} & 
    \multicolumn{1}{|c|}{9.20} & 
    \multicolumn{1}{c|}{\textbf{3.84}} & 
    \multicolumn{1}{c|}{14.39} & 
    \multicolumn{1}{c|}{\textbf{7.23}}
    \\ 
    \hline
    \multicolumn{1}{|c}{10} & 
    \multicolumn{1}{|c|}{8.00} & 
    \multicolumn{1}{c|}{\textbf{3.15}} & 
    \multicolumn{1}{c|}{13.16} & 
    \multicolumn{1}{c|}{\textbf{6.26}} \\ 
    \hline
\end{tabular}
    \label{table:TransformersComparisonTypeNet}
\end{table}

Table \ref{table:TransformersComparisonTypeNet} provides the system performance results in terms of EER (\%) of the proposed Transformer for the different number of enrolment sessions ($E = 1, 5, 10$) and the two different threshold configurations, i.e., computing an individual threshold per subject and obtaining the mean EER over all subjects (Average EER), and a unique threshold for all subjects (Global EER). In addition, to provide a better comparison of the proposed Transformer with recent state-of-the-art keystroke biometric systems, we include the results achieved by TypeNet \cite{acien2021typenet}. TypeNet is based on a  Long Short-Term Memory (LSTM) RNN architecture, achieving state-of-the-art performance results in both physical and touchscreen keyboards. Several learning approaches were also studied using different loss functions (softmax, contrastive, and triplet loss). 
It is important to highlight that we opt for the comparison with TypeNet as \textit{(i)} they considered one of the largest mobile free-text keystroke databases available up to date, the Aalto mobile keystroke database \cite{palin2019people}, \textit{(ii)} their experimental protocol is publicly available in GitHub\footnote{\texttt{\url{https://github.com/BiDAlab/TypeNet}}}, so we can rigorously follow it, considering the same sets of subjects and metrics, for development and evaluation, and \textit{(iii)} TypeNet has outperformed previous approaches in keystroke biometrics.  

As can be seen in Table \ref{table:TransformersComparisonTypeNet}, the proposed Transformer significantly outperforms TypeNet in all cases on the same evaluation set of 1,000 subjects from the Aalto mobile keystroke database \cite{palin2019people}. Analysing the number of enrolment sessions, the proposed Transformer achieves a 6.99\% EER when considering $E=1$ single enrolment session. This is an absolute improvement of 5.61\% EER compared with TypeNet (12.60\% EER), proving the high potential of the proposed Transformer compared with traditional deep learning architectures such as RNNs. Increasing the number of enrolment sessions per subject, we can see a general improvement of the proposed Transformer, with values of 3.84\% and 3.15\% EERs for $E = 5, 10$ enrolment sessions, respectively. This trend also shows the large improvement of the proposed Transformer with the number of enrolment sessions, outperforming TypeNet in both scenarios (absolute improvement of around 5\% EER).

Analysing the two different threshold scenarios (Average and Global), we can observe better results for the Average case regardless of the number of enrolment sessions, e.g., for $E=1$ single enrolment session, values of 6.99\% and 10.68\% EERs are achieved by the Proposed Transformer for the Average and Global cases, respectively. A similar trend is observed for TypeNet. These results make sense as one specific threshold is adapted to each subject in the Average case, at the expense of more computational efforts.   

\begin{figure}[tp]
    \centering
     \includegraphics[trim={0cm 0cm 0 0cm}, width=.92\linewidth]{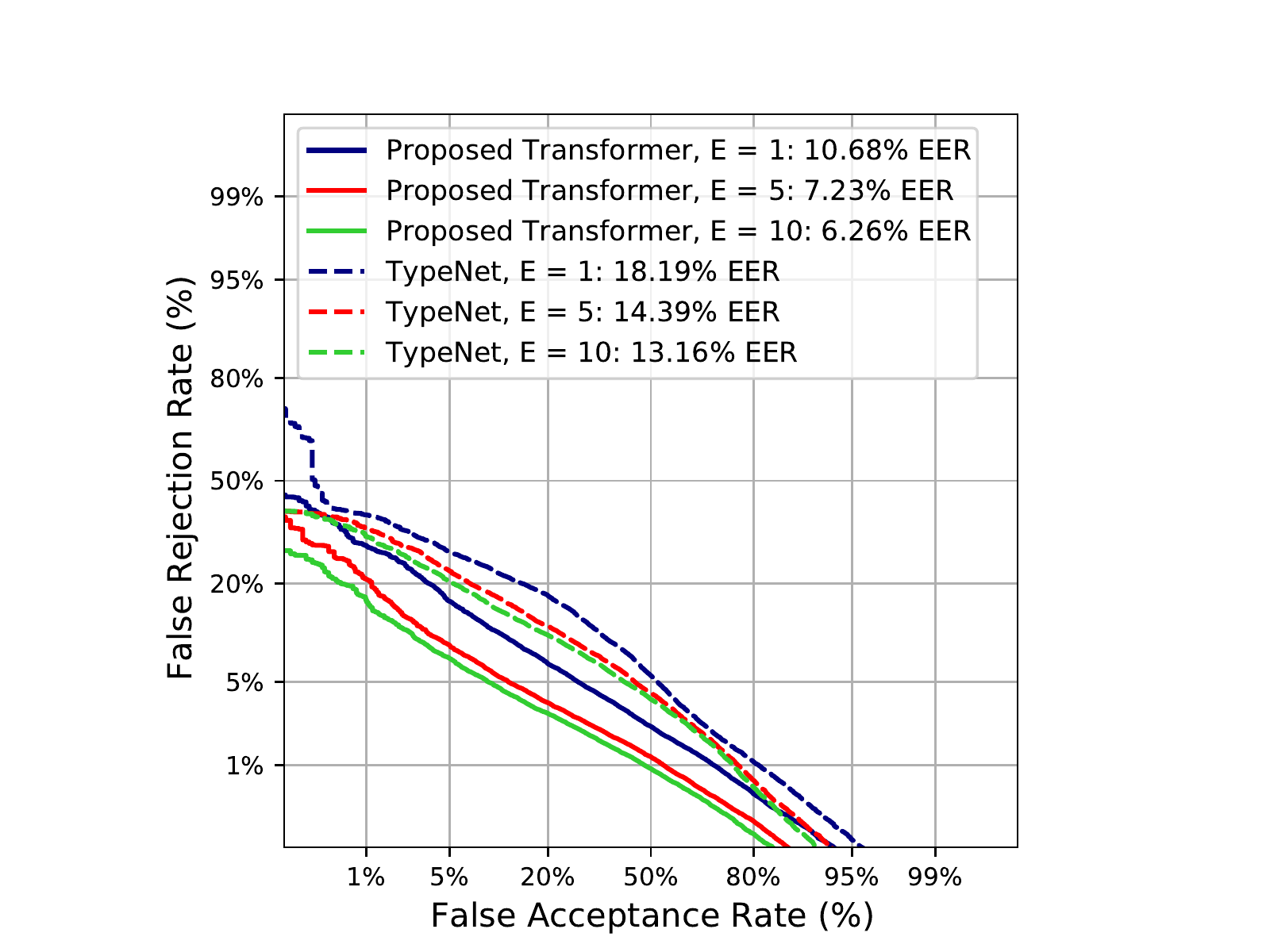}
    \caption{DET curves comparing the performance of the proposed Transformer with TypeNet \cite{acien2021typenet}. \textit{E} corresponds to the number of enrolment sessions considered. The reported EERs (\%) are for the global threshold.}
    \label{fig:dets}
\end{figure}

For completeness, we include in Fig. \ref{fig:dets} the Detection Error Trade-off (DET) curves computed for the different number of enrolment sessions available for the Global EER threshold case. As can be seen, the proposed Transformer outperforms TypeNet even in the case of having fewer enrolment sessions available, i.e., $E=1$ enrolment session (proposed Transformer achieves 10.68\% EER) vs. $E=10$ (TypeNet achieves 13.16\% EER). 

\begin{figure}[th!]
    \centering
     \includegraphics[trim={0cm 0cm 0 0cm}, width=0.92\linewidth]{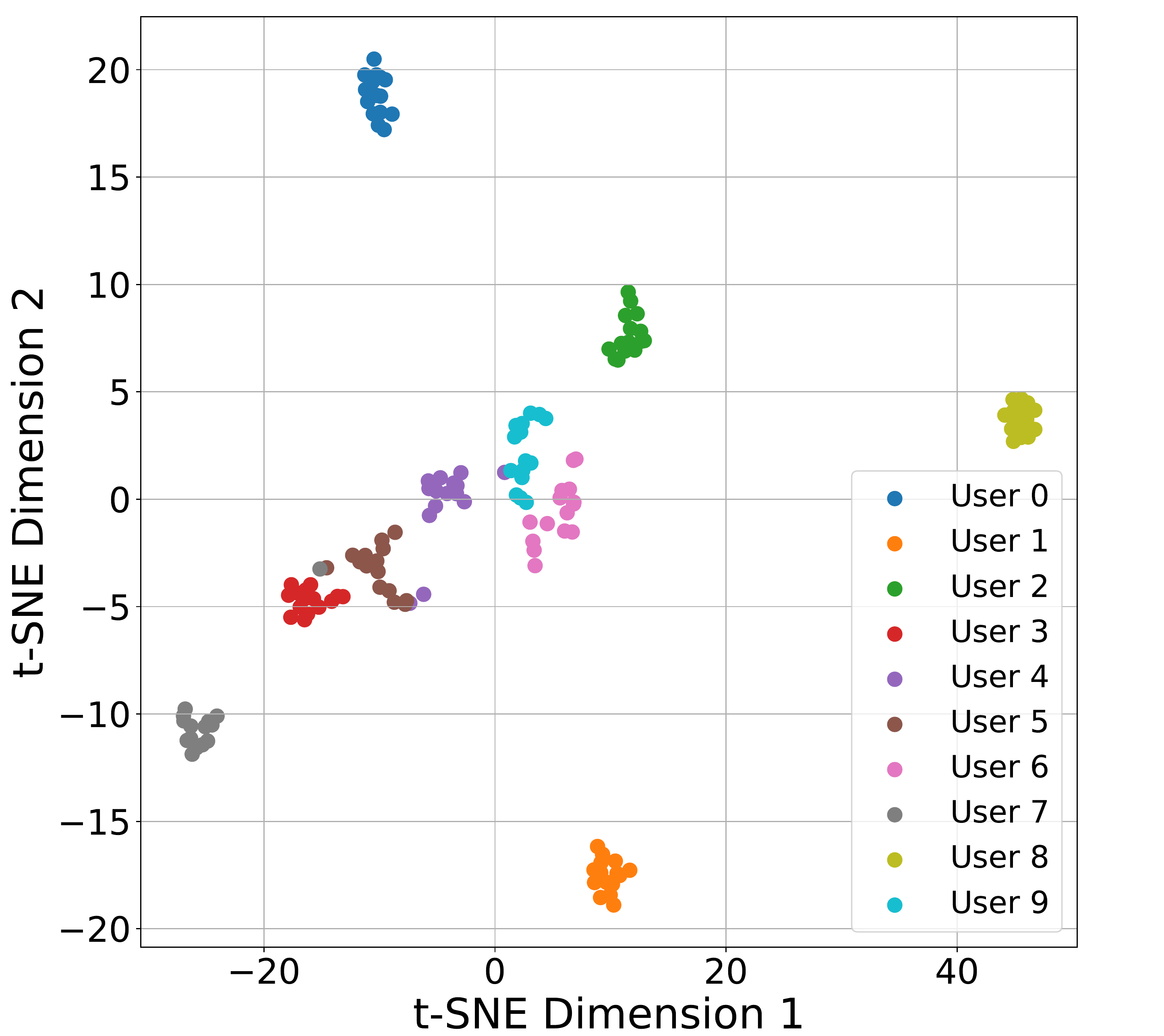}
  \caption[]{2D graphical visualisation of the latent space through t-SNE considering 15 sessions of 10 subjects \cite{van2008visualizing}. Selected parameters\footnotemark: $\texttt{perplexity} = 14$, $\texttt{init} = \texttt{'pca'}$, $\texttt{n\_iter} = 1000$.}
    \label{fig:t-sne}
\end{figure}

\begin{table}[tp!]
    \centering
    \caption{Comparison of the performance achieved by the proposed Transformer with related systems  ($E = 5$).}
    \begin{tabular}{|c|c|}
\hline
\textbf{System}               & \textbf{Average EER {(}\%{)}} \\ \hline
POHMM \cite{monaco2018partially}           & 40.40                   \\ \hline
Digraphs \cite{cceker2016user}        & 29.20                   \\ \hline
CNN+RNN  \cite{lu2020continuous}            & 12.20                   \\ \hline
TypeNet   \cite{acien2021typenet}           & 9.20                    \\ \hline
\textbf{Proposed Transformer} & \textbf{3.84}                    \\ \hline
\end{tabular}
    \label{table:TransformersComparisonSOTA}
\end{table}

Finally, to provide a better comparison of the proposed Transformer with the literature, we include in Table \ref{table:TransformersComparisonSOTA} the EER results obtained by other state-of-the-art systems in keystroke biometrics: digraphs and SVM \cite{cceker2016user}, Partially Observable Hidden Markov Model (POHMM) \cite{monaco2018partially}, and a combination of RNNs and CNNs \cite{lu2020continuous}. All of them are trained under the same experimental protocol and evaluated on the same set of 1,000 subjects in terms of Average EER considering $E=5$ enrolment sessions. Our proposed Transformer outperforms all previous approaches with EER absolute improvements of 36.56\% (POHMM \cite{monaco2018partially}), 25.36\% (Diagraphs \cite{cceker2016user}), 8.36\% (CNN + RNN \cite{lu2020continuous}), and 5.36\% (TypeNet \cite{acien2021typenet}). These results evidence the success and potential of the proposed Transformer for the challenging free-text mobile scenario considered in the study. 

\footnotetext{\texttt{\href{https://scikit-learn.org/stable/modules/generated/sklearn.manifold.TSNE.html}{sklearn.manifold.TSNE -- scikit-learn 1.1.1 documentation}}. Accessed: 2022-07-13.}

\subsection{Analysis of the feature embeddings}
\par Fig. \ref{fig:t-sne} provides a graphical representation of the feature embedding space achieved with the proposed Transformer for 10 different subjects of the Aalto mobile keystroke database (15 acquisition sessions per subject). We consider the popular mathematical method t-SNE \cite{van2008visualizing} to visualise data points in high dimensional spaces. Apart from few outliers (one sample of users 4 and 7), most of the embeddings of each of the subjects are clearly separated. Fig. \ref{fig:t-sne} demonstrates how the proposed Transformer is able to group clearly the feature embeddings belonging to the same subject, achieving small intra-class variability, and to distance as much as possible between the feature embeddings of the different subjects, increasing the inter-class variability.



\subsection{Discussion}
\label{subsec:Discussion}
The system performance improvement achieved with our proposed Transformer in relation with previous approaches is due, in our opinion, to the following reasons: \textit{(i)} our model applies the self-attention mechanism, being able to operate over long-distances in the input sequence; \textit{(ii)} our model attends to all the prior samples of the time sequence at the same time, without summarising the previous seen information; \textit{(iii)} the features are extracted from two different perspectives, from the time and the channel modules, providing more complex information; and \textit{(iv)} the Gaussian range encoding together with the multi-scale keystroke CNN allow to obtain a perspective of each sample in different environments, as different ranges are treated at the same time.

\section{Conclusions and Future Work}
\label{sec:conclusions}

The present study has explored and proposed novel keystroke verification systems based on Transformers. To the best of our knowledge, this is the first attempt to apply Transformers to keystroke biometrics. 
We have focused on the task of free-text mobile keystroke authentication, traditionally far more challenging than its fixed-text counterpart. Our proposed Transformer has greatly reduced the performance gap existing between the two scenarios, reaching numbers as low as 3.15\% EER with 10 short enrolment sessions of 50 samples each, and 3.93\% EER with only 1 enrolment session. Furthermore, for the popular and public Aalto mobile keystroke database considered in the study, the proposed Transformer has achieved remarkable improvements with the same experimental protocol considered in the recently state-of-the-art TypeNet \cite{acien2021typenet} (3.15\% EER vs. 8.00\% EER). 
Finally, it is important to remark that we will make our proposed approach and experimental framework available to the research community in order to advance the state of the art of keystroke biometrics in free-text mobile scenarios\footnote{\texttt{\url{https://github.com/BiDAlab/TypeFormer}}}.

\par Future work will be oriented in several directions: \textit{(i)} improvement of the Transformer architecture; \textit{(ii)} an optimised training approach considering hard triplet mining. Forcing the model to learn from harder comparisons has in fact proved to be an effective strategy in many applications \cite{Schroff_2015_CVPR}; \textit{(iii)} a more sophisticated mechanism than the traditional Euclidean distance for the comparison of feature embeddings in the latent space, such as Support Vector Machines (SVM); \textit{(iv)} investigating the subject information contained in the feature embeddings, i.e., gender, age, etc., to determine if keystroke data should be treated as privacy-sensitive biometric data. The metadata available in the Aalto mobile keystroke database can be used to shed some light; and \textit{(v)} applying Transformers to other biometric modalities \cite{TOLOSANA2022108609}.

\newpage

\section{Acknowledgements}
This project has received funding from the European Union’s Horizon 2020 research and innovation programme under the Marie Skłodowska-Curie grant agreement no. 860315. R. Tolosana and R. Vera-Rodriguez are also supported by INTER-ACTION (PID2021-126521OB-I00 MICINN/FEDER).

{\small
\bibliographystyle{ieee}
\bibliography{0_Main}

\begin{thebibliography}{10}\itemsep=-1pt

\bibitem{bbreport}
{Behavioral Biometrics Market Size \& Share Report, 2020-2027}.
\newblock
  \url{https://www.grandviewresearch.com/industry-analysis/behavioral-biometric-market}.
\newblock Accessed: 2022-07-11.

\bibitem{ISO2018}
{\em {ISO} 9241-11:2018(en): Ergonomics of Human-System Interaction}, 2018.
\newblock {Part 11: Usability: Definitions and Concepts}.

\bibitem{Abuhamad2021}
M.~Abuhamad, A.~Abusnaina, D.~Nyang, and D.~Mohaisen.
\newblock {Sensor-Based Continuous Authentication of Smartphones’ Users Using
  Behavioral Biometrics: A Contemporary Survey}.
\newblock {\em IEEE Internet of Things Journal}, 2021.

\bibitem{acien2021typenet}
A.~Acien, A.~Morales, J.~V. Monaco, R.~Vera-Rodriguez, and J.~Fierrez.
\newblock {TypeNet: Deep Learning Keystroke Biometrics}.
\newblock {\em IEEE Transactions on Biometrics, Behavior, and Identity
  Science}, 2021.

\bibitem{cceker2016user}
H.~{\c{C}}eker and S.~Upadhyaya.
\newblock {User Authentication with Keystroke Dynamics in Long-text Data}.
\newblock In {\em Proc. IEEE Int.l Conf. on Biometrics Theory, Applications and
  Systems}, 2016.

\bibitem{delgadosantos2021survey}
P.~Delgado-Santos, G.~Stragapede, R.~Tolosana, R.~Guest, F.~Deravi, and
  R.~Vera-Rodriguez.
\newblock {A Survey of Privacy Vulnerabilities of Mobile Device Sensors}.
\newblock {\em ACM Comput. Surv.}, 2022.

\bibitem{gigaword}
D.~Graff and C.~Cieri.
\newblock {English Gigaword LDC2003T05}.
\newblock {\em Philadelphia: Linguistic Data Consortium}, 2003.

\bibitem{li2021two}
B.~Li, W.~Cui, W.~Wang, L.~Zhang, Z.~Chen, and M.~Wu.
\newblock {Two-stream Convolution Augmented Transformer for Human Activity
  Recognition}.
\newblock In {\em Proc. AAAI Conf. on Artificial Intelligence}, 2021.

\bibitem{lu2020continuous}
X.~Lu, S.~Zhang, P.~Hui, and P.~Lio.
\newblock {Continuous Authentication by Free-text Keystroke based on CNN and
  RNN}.
\newblock {\em Computers \& Security}, 2020.

\bibitem{Maiorana2021}
E.~Maiorana, H.~Kalita, and P.~Campisi.
\newblock {Mobile Keystroke Dynamics for Biometric Recognition: An Overview}.
\newblock {\em IET Biometrics}, 2021.

\bibitem{monaco2018partially}
J.~V. Monaco and C.~C. Tappert.
\newblock {The Partially Observable Hidden Markov Model and its Application to
  Keystroke Dynamics}.
\newblock {\em Pattern Recognition}, 2018.

\bibitem{mondal2017study}
S.~Mondal and P.~Bours.
\newblock {A Study on Continuous Authentication Using a Combination of
  Keystroke and Mouse Biometrics}.
\newblock {\em Neurocomputing}, 2017.

\bibitem{palin2019people}
K.~Palin, A.~M. Feit, S.~Kim, P.~O. Kristensson, and A.~Oulasvirta.
\newblock {How Do People Type on Mobile Devices? Observations from a Study with
  37,000 Volunteers}.
\newblock In {\em Proc. Int.l Conf. on Human-Computer Interaction with Mobile},
  2019.

\bibitem{7503170}
V.~M. Patel, R.~Chellappa, D.~Chandra, and B.~Barbello.
\newblock {Continuous User Authentication on Mobile Devices: Recent Progress
  and Remaining Challenges}.
\newblock {\em IEEE Signal Processing Magazine}, 2016.

\bibitem{Schroff_2015_CVPR}
F.~Schroff, D.~Kalenichenko, and J.~Philbin.
\newblock {FaceNet: A Unified Embedding for Face Recognition and Clustering}.
\newblock In {\em Proc. IEEE Conf. on Computer Vision and Pattern Recognition},
  2015.

\bibitem{tay2020efficient}
Y.~Tay, M.~Dehghani, D.~Bahri, and D.~Metzler.
\newblock {Efficient Transformers: A Survey}.
\newblock {\em ACM Comput. Surv.}, 2022.

\bibitem{TEH2016210}
P.~S. Teh, N.~Zhang, A.~B.~J. Teoh, and K.~Chen.
\newblock {A Survey on Touch Dynamics Authentication in Mobile Devices}.
\newblock {\em Computers \& Security}, 2016.

\bibitem{TOLOSANA2022108609}
R.~Tolosana, R.~Vera-Rodriguez, C.~Gonzalez-Garcia, J.~Fierrez, A.~Morales,
  J.~Ortega-Garcia, J.~C. Ruiz-Garcia, S.~Romero-Tapiador, S.~Rengifo,
  M.~Caruana, et~al.
\newblock {SVC-onGoing: Signature Verification Competition}.
\newblock {\em Pattern Recognition}, 127:108609, 2022.

\bibitem{van2008visualizing}
L.~Van~der Maaten and G.~Hinton.
\newblock {Visualizing Data using t-SNE}.
\newblock {\em Journal of Machine Learning Research}, 2008.

\bibitem{vaswani2017attention}
A.~Vaswani, N.~Shazeer, N.~Parmar, J.~Uszkoreit, L.~Jones, A.~N. Gomez,
  L.~Kaiser, and I.~Polosukhin.
\newblock {Attention is All you Need}.
\newblock In {\em Proc. Advances in Neural Information Processing Systems},
  2017.

\bibitem{10.1145/2037373.2037418}
K.~Vertanen and P.~O. Kristensson.
\newblock {A Versatile Dataset for Text Entry Evaluations Based on Genuine
  Mobile Emails}.
\newblock In {\em Proc. Int.l Conf. on Human Computer Interaction with Mobile
  Devices and Services}, 2011.

\end{thebibliography}
}

\end{document}